\documentclass[conference]{IEEEtran}
\IEEEoverridecommandlockouts

\usepackage{cite}
\usepackage{amsmath,amssymb,amsfonts}
\usepackage{algorithmic}
\usepackage{graphicx}
\usepackage{textcomp}
\usepackage{xcolor}
\usepackage{subcaption}
\usepackage{siunitx}
\usepackage{booktabs}
\usepackage{multirow}
\usepackage{bm}
\usepackage{arydshln}
\usepackage{graphbox}
\usepackage{paralist}

\usepackage[T1]{fontenc}
\usepackage[hidelinks]{hyperref}

\DeclareMathOperator*{\argmax}{argmax} 

\def\BibTeX{{\rm B\kern-.05em{\sc i\kern-.025em b}\kern-.08em
    T\kern-.1667em\lower.7ex\hbox{E}\kern-.125emX}}

\begin{document}

\title{UAV Path Planning for Wireless Data Harvesting: A Deep Reinforcement Learning Approach\\
\thanks{H. Bayerlein and D. Gesbert are supported by the PERFUME project funded by the European Research Council (ERC) under the European Union's Horizon 2020 research and innovation program (grant agreement no. 670896). M. Caccamo was supported by an Alexander von Humboldt Professorship endowed by the German Federal Ministry of Education and Research. The code for this work is available under \href{https://github.com/hbayerlein/uav_data_harvesting}{https://github.com/hbayerlein/uav\_data\_harvesting}.}
}

\author{\IEEEauthorblockN{Harald Bayerlein$^{1}$, Mirco Theile$^{2}$, Marco Caccamo$^{2}$, and David Gesbert$^{1}$}
\IEEEauthorblockA{$^{1}$Communication Systems Department, EURECOM, Sophia Antipolis, France\\
$^{2}$TUM Department of Mechanical Engineering, Technical University of Munich, Germany\\
\text{\{harald.bayerlein, david.gesbert\}@eurecom.fr, \{mirco.theile, mcaccamo\}@tum.de}\\
}
}

\maketitle

\begin{abstract}
Autonomous deployment of unmanned aerial vehicles (UAVs) supporting next-generation communication networks requires efficient trajectory planning methods. We propose a new end-to-end reinforcement learning (RL) approach to UAV-enabled data collection from Internet of Things (IoT) devices in an urban environment. An autonomous drone is tasked with gathering data from distributed sensor nodes subject to limited flying time and obstacle avoidance. While previous approaches, learning and non-learning based, must perform expensive recomputations or relearn a behavior when important scenario parameters such as the number of sensors, sensor positions, or maximum flying time, change, we train a double deep Q-network (DDQN) with combined experience replay to learn a UAV control policy that generalizes over changing scenario parameters. By exploiting a multi-layer map of the environment fed through convolutional network layers to the agent, we show that our proposed network architecture enables the agent to make movement decisions for a variety of scenario parameters that balance the data collection goal with flight time efficiency and safety constraints. Considerable advantages in learning efficiency from using a map centered on the UAV's position over a non-centered map are also illustrated. 
\end{abstract}

\section{Introduction}

While unmanned aerial vehicles (UAVs) are envisioned for a multitude of applications, their prospective roles in telecommunications can be classified into two categories: cellular-connected UAVs attached to mobile network links or UAVs providing communication services themselves, e.g. collecting data from distributed Internet of Things (IoT) devices \cite{Zeng2019}. As an example in the context of infrastructure maintenance and preserving structural integrity, Hitachi is already commercially deploying partially autonomous UAVs that collect data from IoT sensors embedded in large infrastructure structures \cite{Minevich2020}.


Collecting data from sensor devices in an urban environment imposes challenging constraints on the trajectory design for autonomous UAVs. Battery energy density restricts mission duration for quadcopter drones severely, while the complex urban environment poses challenges in obstacle avoidance and the adherence to regulatory no-fly zones (NFZs). Additionally, the wireless communication channel is characterized by frequent fluctuations in attenuation through alternating line-of-sight (LoS) and non-line-of-sight (NLoS) links. Deep reinforcement learning (DRL) offers the opportunity to balance challenges and data collection goal for complex environments in a straightforward way by combining them in the reward function. This advantage also holds for other instances of UAV path planning, such as coverage path planning, a classical robotics problem where the UAV's goal is to cover all points inside an area of interest. By basing our proposed method on an approach to UAV coverage path planning \cite{Theile2020}, we would like to highlight the connection between these research areas.

A recent tutorial covering the paradigms of cellular-connected UAVs as well as UAV-assisted communications, including trajectory planning for IoT data collection, is given in \cite{Zeng2019}. Bithas \textit{et al.} \cite{Bithas2019} provide a survey on machine learning techniques, including but not limited to reinforcement learning (RL), for various UAV communications scenarios.

Most existing approaches to UAV data collection are not based on RL and only find a solution for one set of scenario parameters at a time. Esrafilian \textit{et al.} \cite{Esrafilian2018} proposed a two-step algorithm to optimize a UAV's trajectory and its scheduling decisions in an urban data collection mission using a combination of dynamic and sequential convex programming. While set in a similar environment, the scenario does not account for NFZs or obstacle avoidance as the drone is assumed to always fly above the highest building. This also holds for the hybrid offline-online optimization approach presented in \cite{You2020}, where a preliminary trajectory is computed before the UAV's start based on a probabilistic LoS channel model and then optimized while the UAV is on its mission in an online fashion.

(Deep) reinforcement learning has been explored in other related UAV communication scenarios. The approach in the simple scenario of \cite{Bayerlein2018spawc}, where a UAV base station serves two ground users, is focused on showing the advantages of neural network (NN) over table-based Q-learning, while not making any explicit assumptions about the environment at the price of long training time. Deep deterministic policy gradient (DDPG), an actor-critic RL method, was proposed by Qi \textit{et al.} \cite{Qi2020} to learn a continuous control policy for a UAV providing persistent communications coverage to a group of users in an environment without obstacles. If a critical scenario parameter like the number of users changes, the agent has to undergo computationally expensive retraining.

Some works under the paradigm of mobile crowdsensing, where mobile devices are leveraged to collect data of common interest, have also suggested the use of UAVs for data collection. Liu \textit{et al.} \cite{Liu2019} proposed an RL multi-agent DDPG algorithm collecting data simultaneously with ground and aerial vehicles in an environment with obstacles and charging stations. While their approach also makes use of convolutional processing to exploit a map of the environment, they do not center the map on the agent's position, which we show to be highly beneficial. Furthermore, in contrast to our method, control policies have to be relearned entirely when scenario and environmental parameters change.

If deep RL methods are to be applied in real-world missions, the prohibitively high training data demand poses one of the most severe challenges \cite{DulacArnold2019}. This is exacerbated by the fact that even small changes in the scenario, such as the number of sensor devices typically require complete retraining. By taking varying parameters in the design and training of the neural network model into account, we take a step towards the mitigation of this challenge.

The main contributions of this paper are the following:
\begin{itemize}
    \item Introducing a novel DDQN-based method to control a UAV on an IoT data harvesting mission, maximizing collected data under flying time and navigation constraints without prior information about the wireless channel characteristics;
    \item Showing the considerable increase in learning efficiency for the RL agent when exploiting a centered \mbox{multi-layer} map of the environment;
    \item Learning to effectively adapt to variations in environmental and scenario parameters as the first step to more realistic RL methods in the context of UAV IoT data collection.
\end{itemize}
\section{System Model and Problem Formulation}

\subsection{System Model}
\label{subsec:system}

We consider a square grid world of size $M \times M \in \mathbb{N}^2$ with the UAV collecting data from $K$ static IoT devices. The $k$-th device is located on ground level at $\mathbf{u}_k = [x_k,y_k,0]^{\operatorname{T}} \in \mathbb{R}^3$ with $k \in [1,K]$. The UAV's data collection mission is over at time $T \in \mathbb{N}$, where the time horizon is discretized into equal mission time slots $t \in [0, T]$. The UAV's position is given by $[x(t), y(t), h]^{\operatorname{T}} \in \mathbb{R}^3$ with constant altitude $h$. Its 2D projection on the ground is given by $\mathbf{p}(t) = [x(t), y(t)]^{\operatorname{T}}$. Mission time slots are chosen sufficiently small so that the UAV's velocity $v(t)$ can be considered to remain constant in one time slot. The UAV is limited to moving with constant velocity $V$ or hovering, i.e. $v(t)\in \{0, V\}$ for all $t \in [0, T]$.

As it is expected that the communication channel is subject to faster changes than the UAV's movement, we partition each mission time slot $t \in [0, T]$ into a number of $\delta \in \mathbb{N}$ communication time slots. The communication time index is then $n \in [0, N]$ with $N = \delta T$. The number of communication time slots per mission time slot $\delta$ is chosen sufficiently large so that the UAV's position, which is interpolated linearly between $\mathbf{p}(t)$ and $\mathbf{p}(t+1)$, and the channel gain can be considered constant within one communication time slot. 

Similar to the channel model in \cite{Esrafilian2018}, the communication links between UAV and the $K$ IoT devices are modeled as LoS/NLoS point-to-point channels with log-distance path loss and shadow fading. The information rate at time $n$ for the $k$-th device is given by
\begin{equation}\label{eq:rate}
    R_k(n) = \log_2 \left( 1 + \text{SNR}_k(n) \right),
\end{equation}
where the signal-to-noise ratio (SNR) with transmit power $P_k$, white Gaussian noise power at the receiver $\sigma^2$, UAV-device distance $d_k(n)$, path loss exponent $\alpha_l$ and $\eta_l \sim  \mathcal{N}(0,\,\sigma_l^{2})$ modeled as a Gaussian random variable, is defined as
\begin{equation}
    \text{SNR}_k(n) = \frac{P_k}{\sigma^2} \cdot d_k(n)^{-\alpha_l} \cdot 10^{\eta_l/10}.
    \label{eq:snr}
\end{equation}
Note that the urban environment causes a strong dependence of the propagation parameters on the $l \in \{\text{LoS, NLoS}\}$ condition and that \eqref{eq:snr} is the SNR averaged over small scale fading.

The sensor nodes are served by the UAV in a simple time-division multiple access (TDMA) manner where, in each communication time slot $n \in [0, N]$, the sensor node $k \in [1, K]$ with the highest $\text{SNR}_k(n)$ with remaining data to be uploaded is picked by the scheduling algorithm. The TDMA constraint for the scheduling variable $q_k(n) \in \{0, 1\}$ is given by
\begin{equation}
    \sum_{k=1}^K q_k(n) \leq 1,~ n \in\left[0, N\right].
\end{equation}

The achievable throughput for one mission time slot $t$ is then the sum of the achieved rates of the corresponding communication time slots $n \in [\delta t, \delta (t + 1) - 1]$ over $K$ sensor nodes and given by
\begin{equation} \label{eq:throughput}
    C(t) = \sum_{n=\delta t}^{\delta (t + 1) - 1} \sum_{k=1}^K q_k(n) R_k(n).
\end{equation}

The central goal of the trajectory optimization problem is the maximization of throughput over the whole data collection mission while minimizing flight duration, subject to the constraints of maximum flight time, adherence to NFZs, obstacle avoidance, and safe landing in designated landing areas. We translate this optimization problem into a reward function as part of a Markov decision process, which we solve using deep reinforcement learning.

\subsection{Markov Decision Process}

A Markov decision process (MDP) is defined by the tuple $(\mathcal{S}, \mathcal{A}, R, P)$ with state-space $\mathcal{S}$, action space $\mathcal{A}$ and reward function $R$. We consider a finite horizon MDP with a probabilistic state transition function $P : \mathcal{S} \times \mathcal{A} \times \mathcal{S} \rightarrow \mathbb{R}$. In line with standard MDP convention, the time index $t$ is written in subscript in the following.

The state at mission time $t$ in the grid world of size $M \times M$ is given by $s_t = (\mathbf{D}_t,\mathbf{p}_t, b_t,\mathbf{M}, \mathbf{U})$ and consists of five components:
\begin{itemize}
    \item $\mathbf{D}_t \in \mathbb{R}^{K\times 2}$ represents the initially available and the already collected data for each device;
    \item $\mathbf{p}_t \in \mathbb{R}^2$ is the UAV position projected on the ground;
    \item $b_t \in \mathbb{N}$ is the UAV's remaining flying time;
    \item $\mathbf{M} \in \mathbb{B}^{M\times M \times 3}$ is the map of the physical environment in the Boolean domain $\{0, 1\}$ encoded with three map layers for start/landing positions, NFZs and buildings;
    \item $\mathbf{U} \in \mathbb{R}^{K\times 2}$ are the 2D coordinates of the $K$ IoT devices.
\end{itemize}
Note that the state is transformed before being fed into the agent as detailed in \ref{subsec:input_space}. Considering the five described components, the total size of the state space is
\begin{equation*}
    \mathcal{S} = 
    \underbrace{\mathbb{R}^2}_{\text{Position}}\times
    \underbrace{\mathbb{B}^{M\times M \times 3}}_{\substack{\text{Environment}\\ \text{Map}}}\times 
    \underbrace{\mathbb{R}^{K\times2}}_{\substack{\text{Device}\\ \text{Positions}}}\times
    \underbrace{\mathbb{R}^{K\times2}}_{\substack{\text{Device}\\ \text{Data}}}\times
    \underbrace{ \mathbb{N}}_{\substack{\text{Flying}\\ \text{Time}}},
\end{equation*}

\noindent while the UAV is limited to six actions contained in the action space
\begin{equation*}
    \mathcal{A} = \{\text{north}, \text{east}, \text{south}, \text{west}, \text{hover}, \text{land}\}.
\end{equation*}

The reward function maps state-action pairs to a real-valued reward, i.e. $R : \mathcal{S} \times \mathcal{A} \rightarrow \mathbb{R}$. Representing the mission goals, the reward function consists of the following components:
\begin{itemize}
    \item $r_{data}$ \textit{(positive)} the data collection reward given by the achieved throughput \eqref{eq:throughput} in the current time slot;
    \item $r_{sc}$ \textit{(negative)} safety controller (SC) penalty in case the drone has to be prevented from colliding with a building or entering an NFZ;
    \item $r_{mov}$ \textit{(negative)} constant movement penalty that is applied for every action the UAV takes without completing the mission;
    \item $r_{crash}$ \textit{(negative)} penalty in case the drone's remaining flying time reaches zero without having landed safely in a landing zone.
\end{itemize}

\section{Methodology}

\subsection{Q-Learning}

Q-learning is a model-free RL method \cite{Sutton2018} where a cycle of interaction between an agent and the environment enables the agent to learn and optimize a behavior, i.e. the agent observes state $s_t \in \mathcal{S}$ and performs an action $a_t \in \mathcal{A}$ at time $t$ and the environment subsequently assigns a reward $r(s_t, a_t) \in \mathbb{R}$ to the agent. The cycle restarts with the propagation of the agent to the next state $s_{t+1}$. The agent's goal is to learn a behavior rule, referred to as a policy that maximizes the reward it receives. A probabilistic policy $\pi(a|s)$ is a distribution over actions given the state such that $\pi : \mathcal{S}\times\mathcal{A}\rightarrow\mathbb{R}$. In the deterministic case, it reduces to $\pi(s)$ such that $\pi : \mathcal{S}\rightarrow\mathcal{A}$.

Q-learning is based on iteratively improving the state-action value function or Q-function to guide and evaluate the process of learning a policy $\pi$. It is given as
\begin{equation}
    Q^\pi(s,a) = \mathbb{E}_{\pi} \left[R_t | s_t = s, a_t = a\right]
    \label{eq:q}
\end{equation}
and represents an expectation of the discounted cumulative return $R_t$  from the current state $s_t$ up to a terminal state at time $T$ given by
\begin{equation}
    R_t = \sum_{k=t}^T \gamma^{k-t} r(s_k,a_k)
\end{equation}
with $\gamma \in [0, 1]$ being the discount factor, balancing the importance of immediate and future rewards. For the ease of exposition, $s_t$ and $a_t$ are abbreviated to $s$ and $a$ and $s_{t+1}$ and $a_{t+1}$ to $s^\prime$ and $a^\prime$ in the following.

\subsection{Double Deep Q-learning and Combined Experience Replay}

As demonstrated in \cite{Bayerlein2018spawc}, representing the Q-function \eqref{eq:q} as a table of values is not efficient in the large state and action spaces of UAV trajectory planning. Through the work of Mnih \textit{et al.} \cite{Mnih2015} on the application of techniques such as experience replay, it became possible to stably train large neural networks with parameters $\theta$, referred to as deep Q-networks (DQNs), to approximate the Q-function instead.

Experience replay is a technique to reduce correlations in the sequence of training data where new experiences made by the agent, represented by quadruples of $(s, a, r, s^\prime)$, are stored in the replay memory $\mathcal{D}$. During training, minibatches of size $m$ are sampled uniformly from $\mathcal{D}$, where the buffer size $\mathcal{|\mathcal{D}|}$ was shown to be an important hyperparameter for the agent's performance and must be carefully tuned for different scenarios. Zhang and Sutton \cite{Zhang2017} proposed combined experience replay as a remedy for this sensitivity with very low computational complexity $\mathcal{O}(1)$. Then, only $m-1$ samples of the minibatch are sampled from memory, while the agent's latest experience is always added. Therefore, all new transitions influence the agent immediately, making the agent less sensitive to the selection of the replay buffer size.

Further improvements to the training process were suggested in \cite{VanHasselt2016}, resulting in the inception of double deep Q-networks (DDQNs). We train our network with parameters $\theta$ accordingly to minimize the loss function given by
\begin{equation}
    L(\theta) = \mathbb{E}_{s,a,s^\prime \sim \mathcal{D}}[(Q_\theta(s,a) - Y(s,a,s^\prime))^2]
    \label{eq:loss}
\end{equation}
where the target value, computed using a separate target network with parameters $\bar{\theta}$, is given by 
\begin{equation}
    Y(s,a,s^\prime) = r(s,a) + \gamma Q_{\bar{\theta}}(s^\prime, \argmax_{a^\prime}Q_{\theta}(s^\prime, a^\prime)).
    \label{eq:target}
\end{equation}

\subsection{Centered Global Map}
\label{subsec:input_space}
The global map is composed of the static environmental map and a dynamic device data map, which is formatted as two real-valued map layers. The first layer represents the data available for collection from each device at its respective position and the second layer records the data that has already been collected throughout the mission.

With this encoding, it would be possible to feed the map data directly into the agent as it was done in \cite{Theile2020}, with an input space defined through
\begin{equation*}
    \mathcal{I} = 
    \underbrace{\mathbb{R}^2}_{\text{Position}}\times
    \underbrace{\mathbb{B}^{M\times M \times 3}}_{\substack{\text{Environment}\\ \text{Map}}}\times 
    \underbrace{\mathbb{R}^{M\times M \times 2}}_{\substack{\text{Device Data}\\ \text{Map}}}\times 
    \underbrace{\mathbb{N}}_{\substack{\text{Flying}\\ \text{Time}}}.
\end{equation*}
In this work, we show that centering the map layers on the UAV's position greatly benefits its ability to generalize over varying scenario parameters. While centering an input map was already applied to local maps that only show the area immediately surrounding the agent, such as in the related field of UAV navigation \cite{Maciel2019}, we apply it for the first time to global maps in a UAV data collection scenario. 

\begin{figure}
    \centering
    \begin{subfigure}{0.48\columnwidth}
        \centering
        \includegraphics[width=\textwidth]{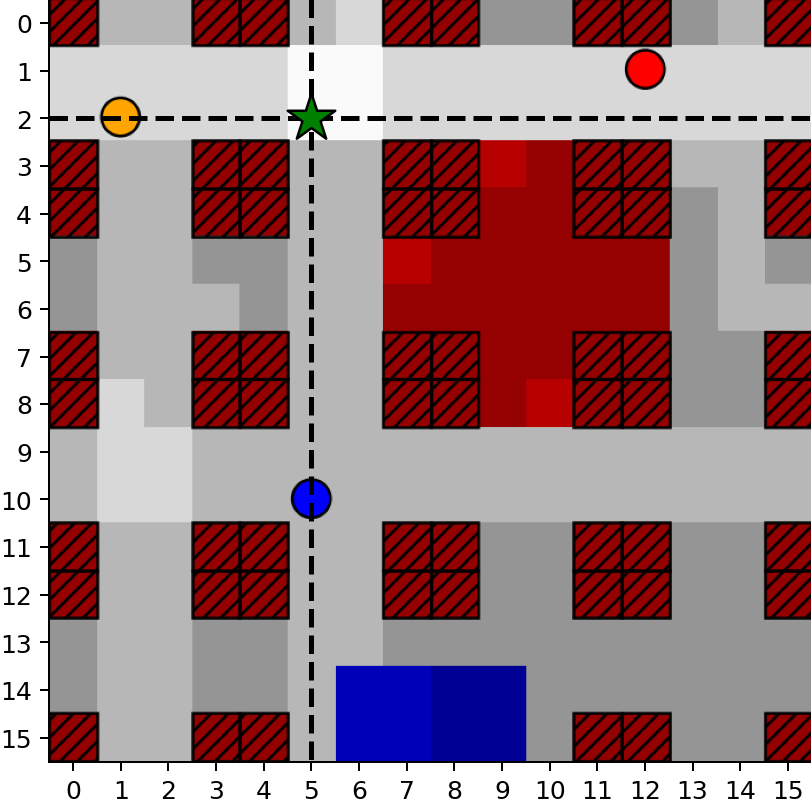}
        \caption{Non-centered input map}
    \end{subfigure}\hspace{5pt}%
    \begin{subfigure}{0.48\columnwidth}
        \centering
        \includegraphics[width=\textwidth]{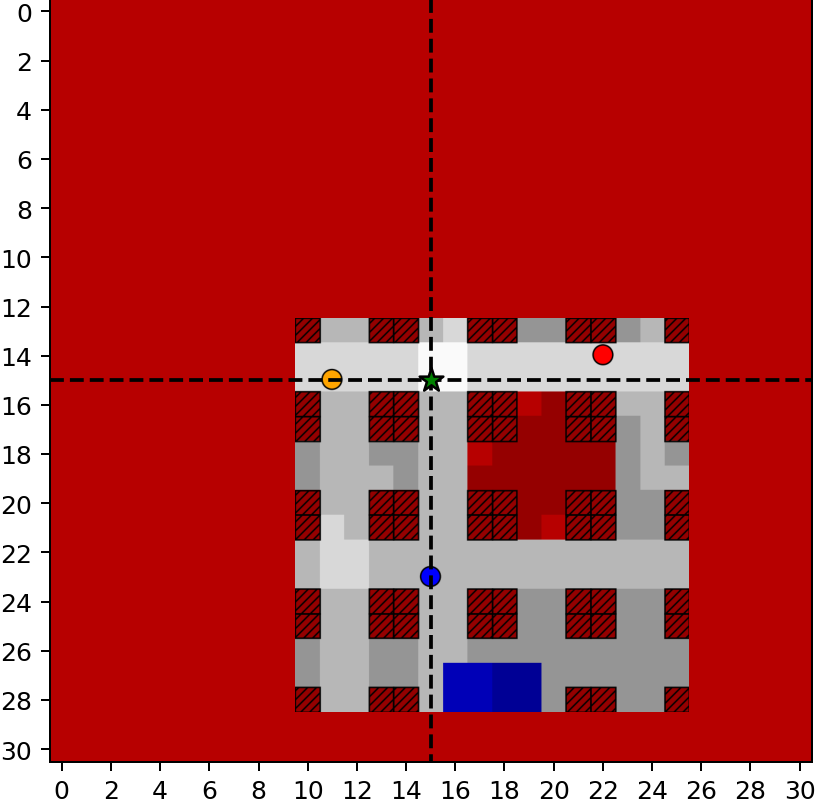}
        \caption{Centered input map}
    \end{subfigure}
    \caption{Comparison of non-centered and centered input maps, with UAV's position represented by the green star and the intersection of the dashed lines.}\vspace{-10pt}
    \label{fig:centering}
\end{figure}

The map centering process inside the computational graph is illustrated in Fig. \ref{fig:centering} with a legend provided in Table \ref{table:legend}. For centering, the maps are expanded to $(2M-1)\times (2M-1)$ in order to enable the agent to observe the entire map independent of its position in it. Translation of the original map centers the expanded map on the UAV's position. The resulting input space is defined through
\begin{equation*}
    \mathcal{I}_c = 
    \underbrace{\mathbb{B}^{(2M-1)\times (2M-1) \times 3}}_{\substack{\text{Centered Environment}\\ \text{Map}}}\times 
    \underbrace{\mathbb{R}^{(2M-1) \times (2M-1)\times 2}}_{\substack{\text{Centered Device}\\ \text{Data Map}}}\times 
    \underbrace{ \mathbb{N}}_{\substack{\text{Flying}\\ \text{Time}}}.
\end{equation*}

The benefit of using a centered map is the result of a change in position to which a neuron of the "Flatten" layer (see Fig. \ref{fig:network}) corresponds. If the map is not centered, the neurons in that layer correspond to features at \textit{absolute} positions. If the map is centered, they correspond to features at positions \textit{relative} to the agent. Since the agent's actions are solely based on its relative position to features, e.g. its distance to devices, learning efficiency increases considerably.

\begin{table}
\vspace{5pt}
\center
\small
\begin{tabular*}{\columnwidth}{lcl}
\toprule[1.5pt]
&Symbol & Description\\
\midrule
\multirow{4}{*}{\rotatebox[origin=c]{90}{\footnotesize{DQN Input}}}
&\includegraphics[align=c,height=.3cm]{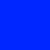} & Start and landing zone\\
&\includegraphics[align=c,height=.3cm]{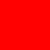} & Regulatory no-fly zone (NFZ)\\
&\includegraphics[align=c,height=.3cm]{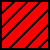} & Buildings blocking wireless links\\
&\includegraphics[align=c,height=.3cm]{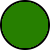} & IoT device\\\midrule
\multirow{5}{*}{\rotatebox[origin=c]{90}{\footnotesize{Visualization}}}
&\includegraphics[align=c,height=.3cm]{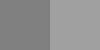} & Summation of building shadows\\
&\includegraphics[align=c,height=.3cm]{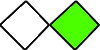} & Starting and landing positions during an episode\\
&\includegraphics[align=c,height=.3cm]{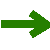} & UAV movement while comm. with  green device\\
&\includegraphics[align=c,height=.3cm]{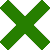} & Hovering while comm. with green device\\
&\includegraphics[align=c,height=.3cm]{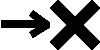} & Actions without comm. (all data collected)\\
\bottomrule[1.5pt]
\end{tabular*}
\caption{Legend for scenario plots.}
\label{table:legend}
\vspace{-5pt}
\end{table}

\subsection{Neural Network Model}
\begin{figure*}
    \centering
    \includegraphics[width=0.9\textwidth]{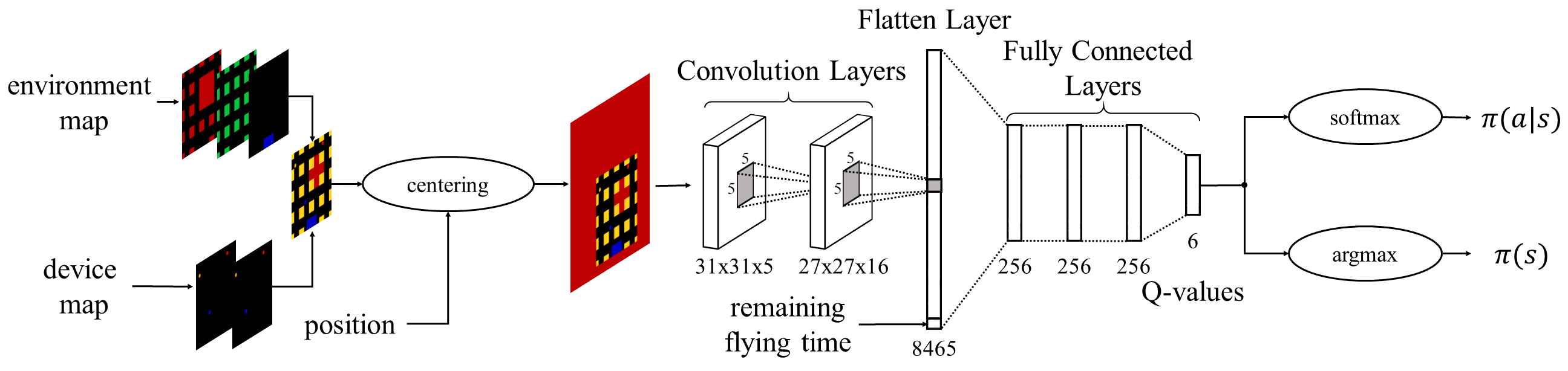}
    \caption{DQN architecture with map centering, with the device map encoded in separate layers but visualized in RGB channels.}
    \label{fig:network}
\end{figure*}

Fig. \ref{fig:network} shows the DQN structure and the map centering pre-processing. The centered map is fed through convolutional layers with ReLU activation and then flattened and concatenated with the scalar input indicating remaining flight time. After passing through fully connected layers with ReLU activation, the data reaches the last fully-connected layer of size $|\mathcal{A}|$ and without activation function, directly representing the Q-values for each action given the input state. The $\argmax$ of the Q-values, the greedy policy is given by
\begin{equation}
    \pi(s) = \argmax_{a\in\mathcal{A}}Q_\theta(s,a).
    \label{eq:greedy}
\end{equation}
It is deterministic and used when evaluating the agent. During training, the soft-max policy 
\begin{equation}
    \pi(a_i|s) = \frac{\mathrm{e}^{Q_\theta(s,a_i)/\beta}}{\sum_{\forall a_j \in \mathcal{A}}\mathrm{e}^{Q_\theta(s,a_j)/\beta}}
    \label{eq:softmax}
\end{equation}
is used. The temperature parameter $\beta \in \mathbb{R}$ scales the balance of exploration versus exploitation. 

\section{Simulations}

\subsection{Simulation Setup}

The UAV starts each new mission in a world discretized into $16 \times 16$ cells where each grid cell is of size $10\si{m} \times 10\si{m}$. It starts with a remaining flying time of $T$ steps, which is decremented by one after every action the agent takes, no matter if moving or hovering. The UAV flies at a constant altitude of $h=10\si{m}$ inside 'urban canyons' through a city environment or open fields and is, for regulatory reasons, not allowed to fly over buildings, enter NFZs, or leave the $16 \times 16$ grid.

Each mission time slot contains $\delta = 4$ scheduled communication time slots. Propagation parameters (see \ref{subsec:system}) are chosen in-line with \cite{Esrafilian2018} according to the urban micro scenario with $\alpha_{\text{LoS}} = 2.27$, $\alpha_{\text{NLoS}} = 3.64$, $\sigma_{\text{LoS}}^2 = 2$ and $\sigma_{\text{NLoS}}^2 = 5$. The shadowing maps to simulate the environment were computed using ray tracing from and to the center points of cells. Transmission and noise powers are normalized through the definition of a cell-edge SNR of -15dB, which describes the SNR between the drone on ground level at the very center of the map and an unobstructed device at one of the grid corners. The agent has absolutely no prior knowledge of the shadowing maps or wireless channel characteristics.

We use the following metrics to evaluate the agent's performance in different scenarios and to compare training instances:
\begin{itemize}
    \item \textit{Cumulative reward}: the sum of all rewards received throughout an episode;
    \item \textit{Has landed}: records whether the agent landed in time at the end of an episode;
    \item \textit{Collection ratio}: the ratio of collected data to total initially available device data at the end of a mission;
    \item \textit{Collection ratio and landed}: the product of \textit{has landed} and \textit{collection ratio} per episode.
\end{itemize}

Evaluation is challenging as we train a single agent to generalize over a large scenario parameter space. During training, we evaluate the agent's training progress in a randomly selected scenario every ten episodes and form an average over multiple evaluations. As it is computationally infeasible to evaluate the trained agent on all possible scenario variations, we perform Monte Carlo analysis on a large number of randomly selected scenario parameter combinations.

\subsection{Centered vs. Non-Centered Map}

\begin{figure}
    \centering\vspace{-15pt}
    \begin{subfigure}{0.49\columnwidth}
        \includegraphics[width=\textwidth]{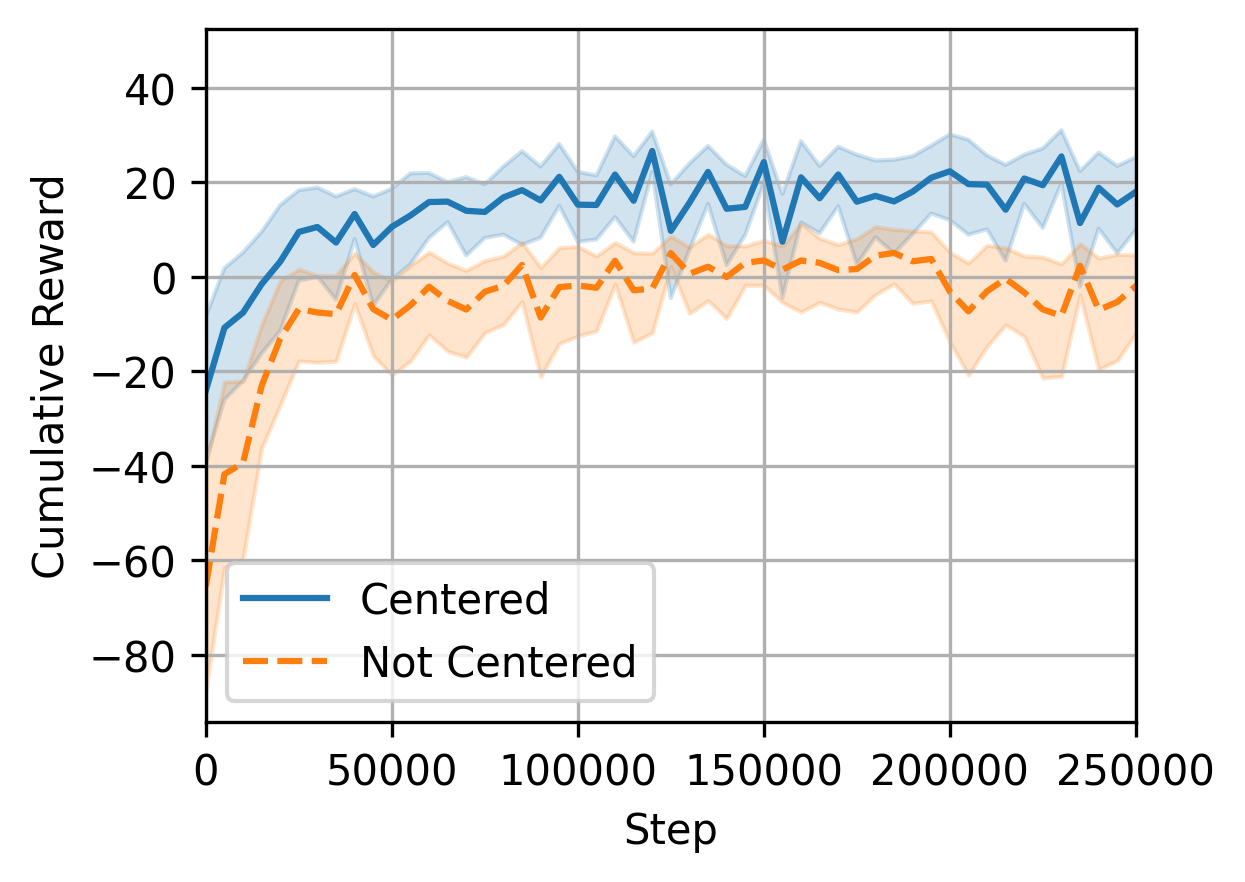}
        \caption{Episodic cumulative reward}
    \end{subfigure}\hspace{5pt}%
    \begin{subfigure}{0.49\columnwidth}
        \includegraphics[width=\textwidth]{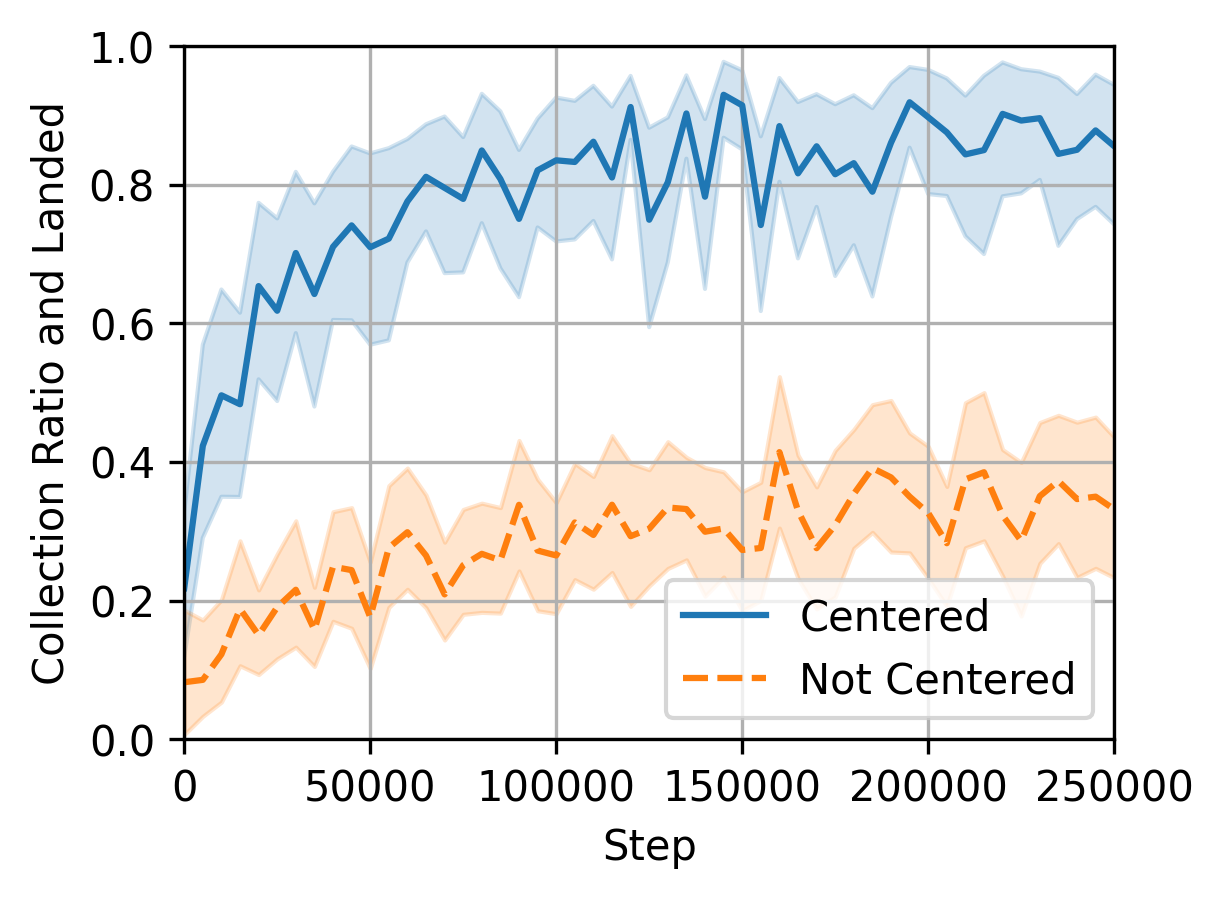}
        \caption{Collection ratio and landed}
    \end{subfigure}
    \caption{Training process comparison between centered and non-centered map input showing the average and 99\% quantiles of three training processes each, with episodic metrics grouped in bins of 5000 step width.}\vspace{-5pt}
    \label{fig:cvnc}
\end{figure}

Centering the map information on the UAV's position as described in \ref{subsec:input_space} proved to be highly beneficial to the learning performance and the generalization ability of the DDQN agent. Fig. \ref{fig:cvnc} shows comparisons of two performance metrics, cumulative reward per episode, and achieved data collection ratio in missions with in-time landing over training time, for centered and \mbox{non-centered} map inputs in identical scenarios. 

To compare the two approaches, the input of the non-centered agent is padded with NFZ cells to have the same shape as the centered agent. The only difference between the agents is that the non-centered agent receives the position as a 2D-one-hot encoded map layer similar to \cite{Theile2020}.
Each graph is averaged over three training runs to account for possible random variations in the training process. A clear performance advantage for the agent using the centered map input can be seen throughout the whole learning process. 

\subsection{Collectible Data and Device Accessibility}

\begin{figure}
    \centering\vspace{-11pt}
    \begin{subfigure}{0.47\columnwidth}
        \includegraphics[width=\textwidth]{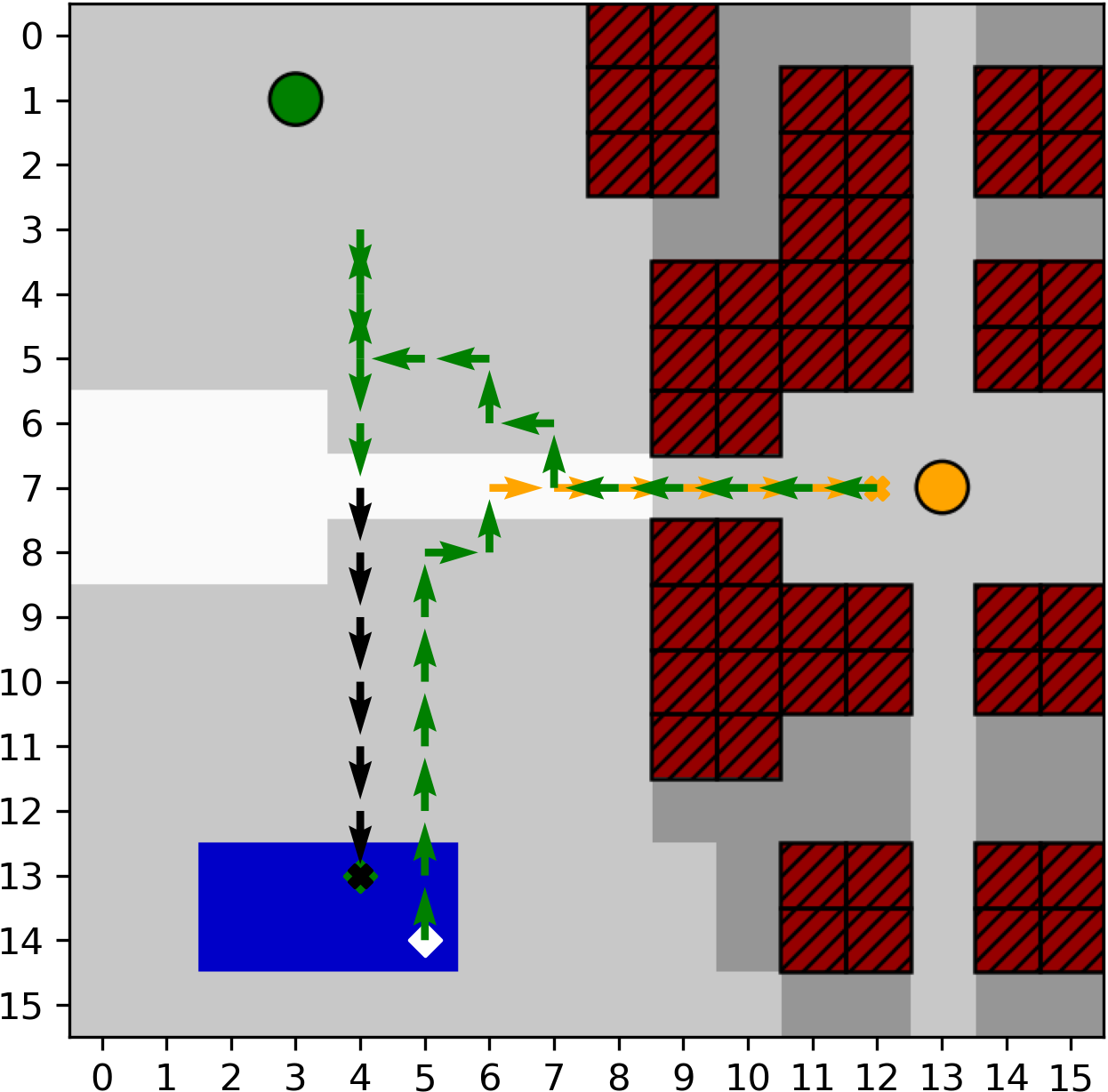}
        \caption{Equal data amounts\newline}
        \label{fig:ip:a}
    \end{subfigure}\hspace{5pt}%
    \begin{subfigure}{0.47\columnwidth}
        \includegraphics[width=\textwidth]{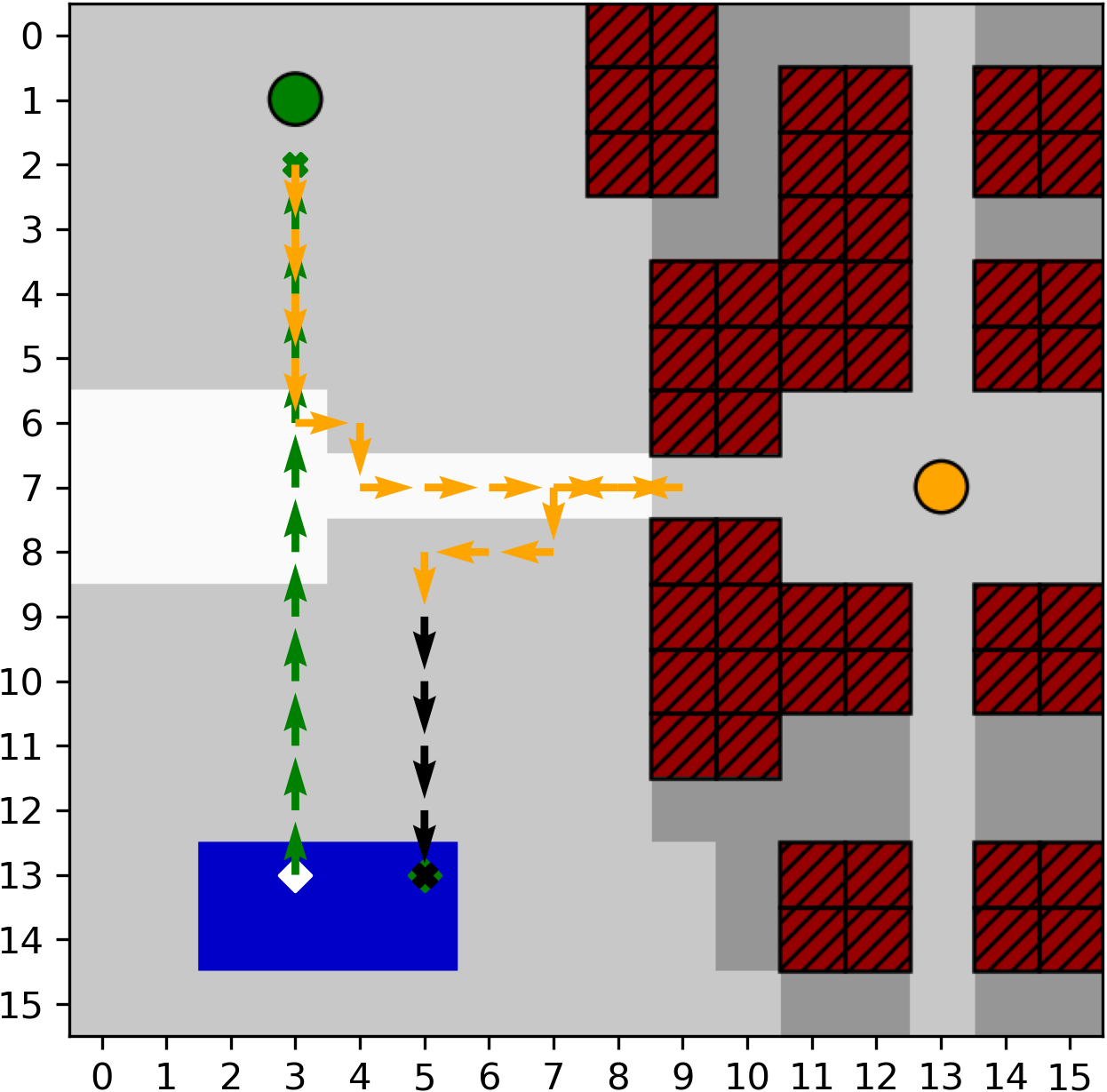}
        \caption{Orange device with a quarter of the green device's data}
        \label{fig:ip:b}
    \end{subfigure}
    \caption{Illustration of the same agent adapting to differences in collectible data with all other mission parameters fixed.}\vspace{-5pt}
    \label{fig:ip}
\end{figure}

The scenario map in Fig. \ref{fig:ip} is divided into an open field and an adjacent city. To show the agent's responsiveness to differences in collectible data at the same devices, we fixed the number of IoT devices to $K=2$, while allowing for fully randomized device positions in unoccupied map space, for each device randomized collectible data ($D_0 \in [1.0, 25.0]$ data units), randomized flying time limits ($b_0 \in [35, 70]$ steps) and eight possible start positions. 

Fig. \ref{fig:ip} shows the agent adapting to a change in collectible data at the two devices. The agent only enters the hard to navigate courtyard if the amount of data at the orange device requires it. While starting to communicate with the unobstructed green device in Fig. \ref{fig:ip:a}, the agent proceeds to collect data from the harder-to-access orange device first, then picking up the rest from the green device before returning straight to the landing area. For the case in Fig. \ref{fig:ip:b}, the UAV changes its strategy. While immediately reducing its distance to the green node after starting and collecting all its data, it collects the data from the orange device on the way back with a detour only as long as required, minimizing the overall mission duration. The UAV is also clearly able to identify unobstructed positions to communicate with the orange device.

\subsection{Manhattan Scenario}

\begin{figure}[h]
    \centering
    \begin{subfigure}{0.47\columnwidth}
        \centering
        \includegraphics[width=\textwidth]{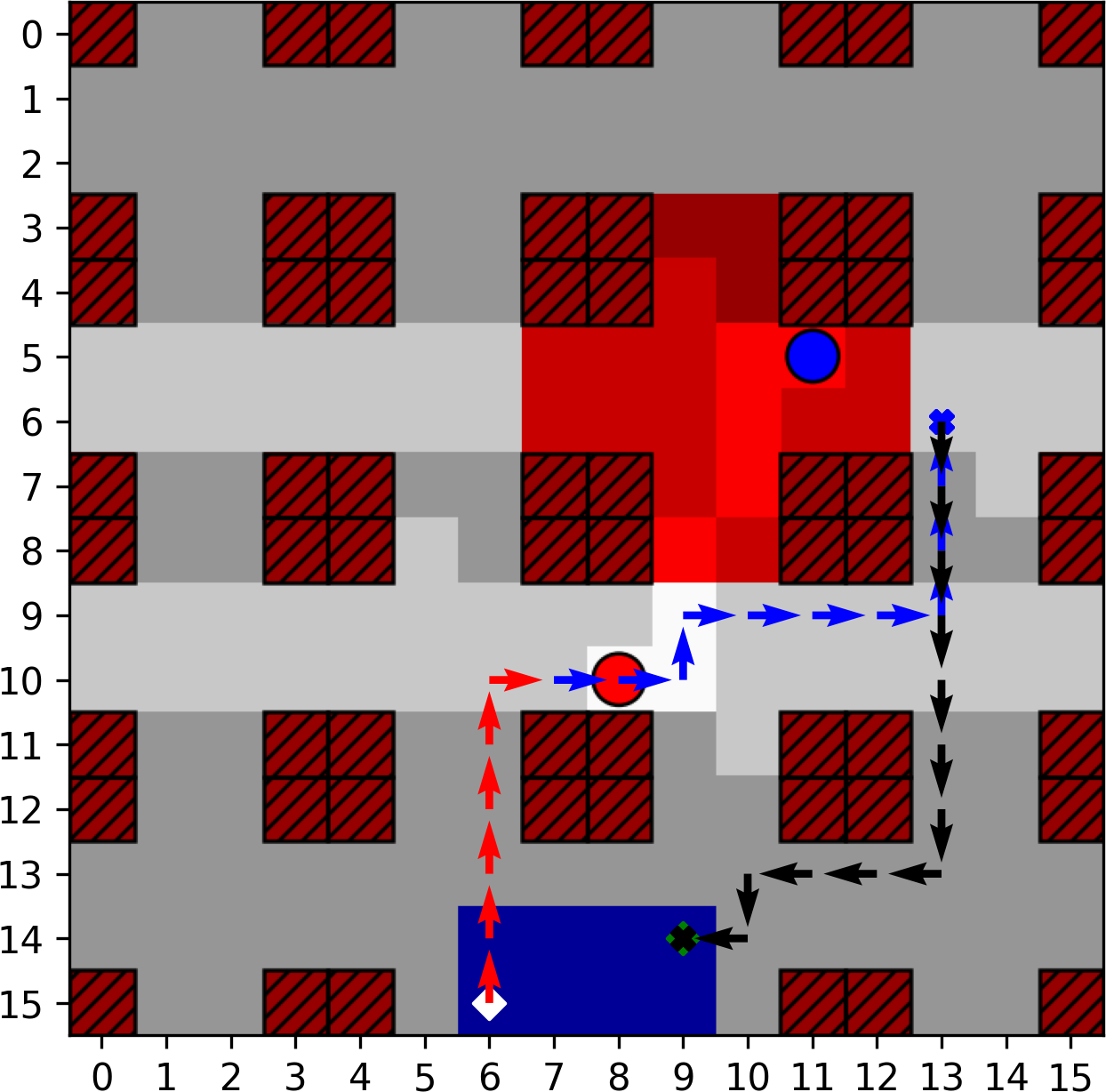}
        \caption{Time 31/38; Data 17.4/17.4}
        \label{fig:mh:a}
    \end{subfigure}\hspace{5pt}%
    \begin{subfigure}{0.47\columnwidth}
        \centering
        \includegraphics[width=\textwidth]{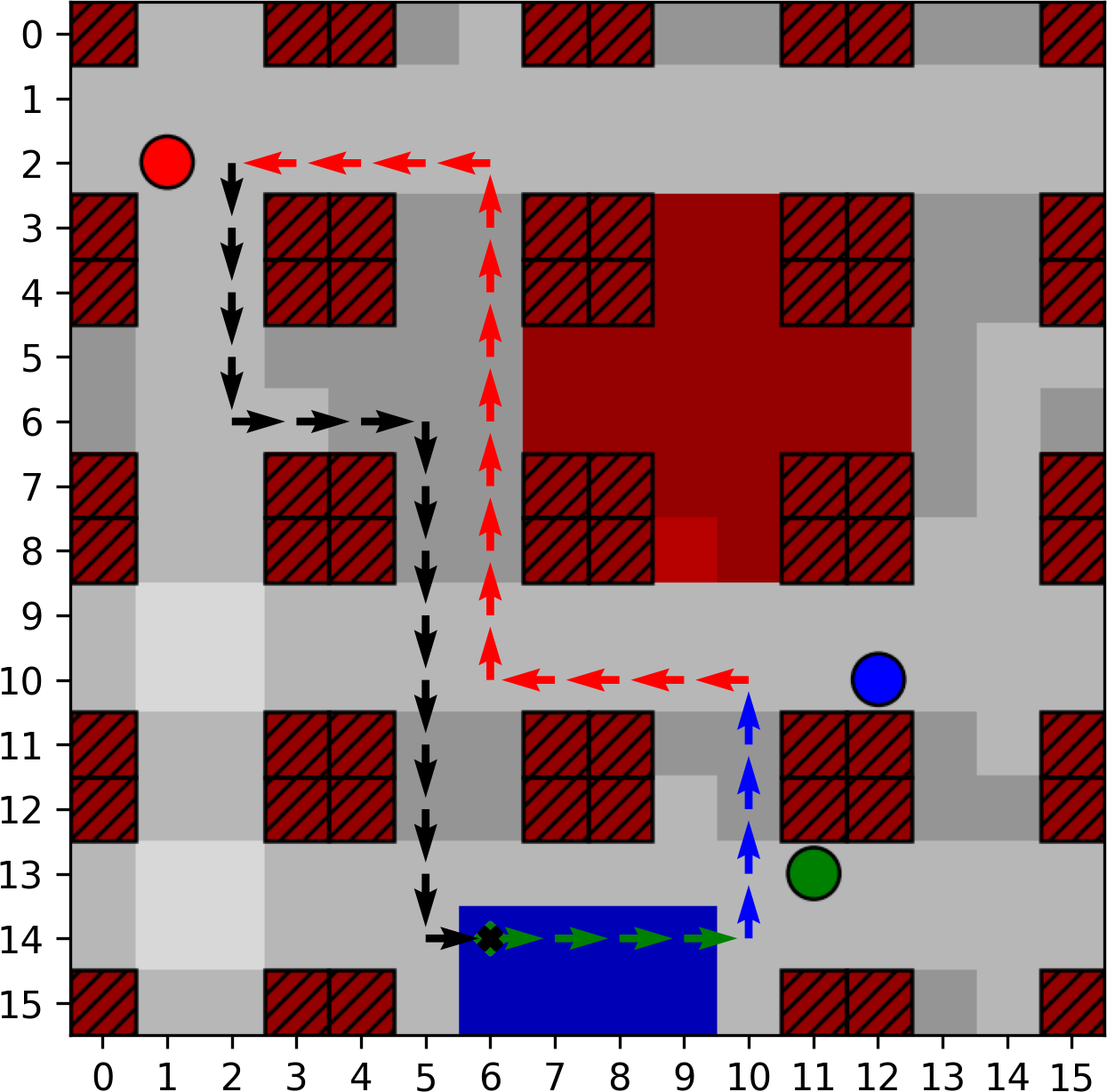}
        \caption{Time 41/41; Data 34.5/34.5}
        \label{fig:mh:b}
    \end{subfigure}
    \begin{subfigure}{0.47\columnwidth}
        \centering
        \vspace{5pt}
        \includegraphics[width=\textwidth]{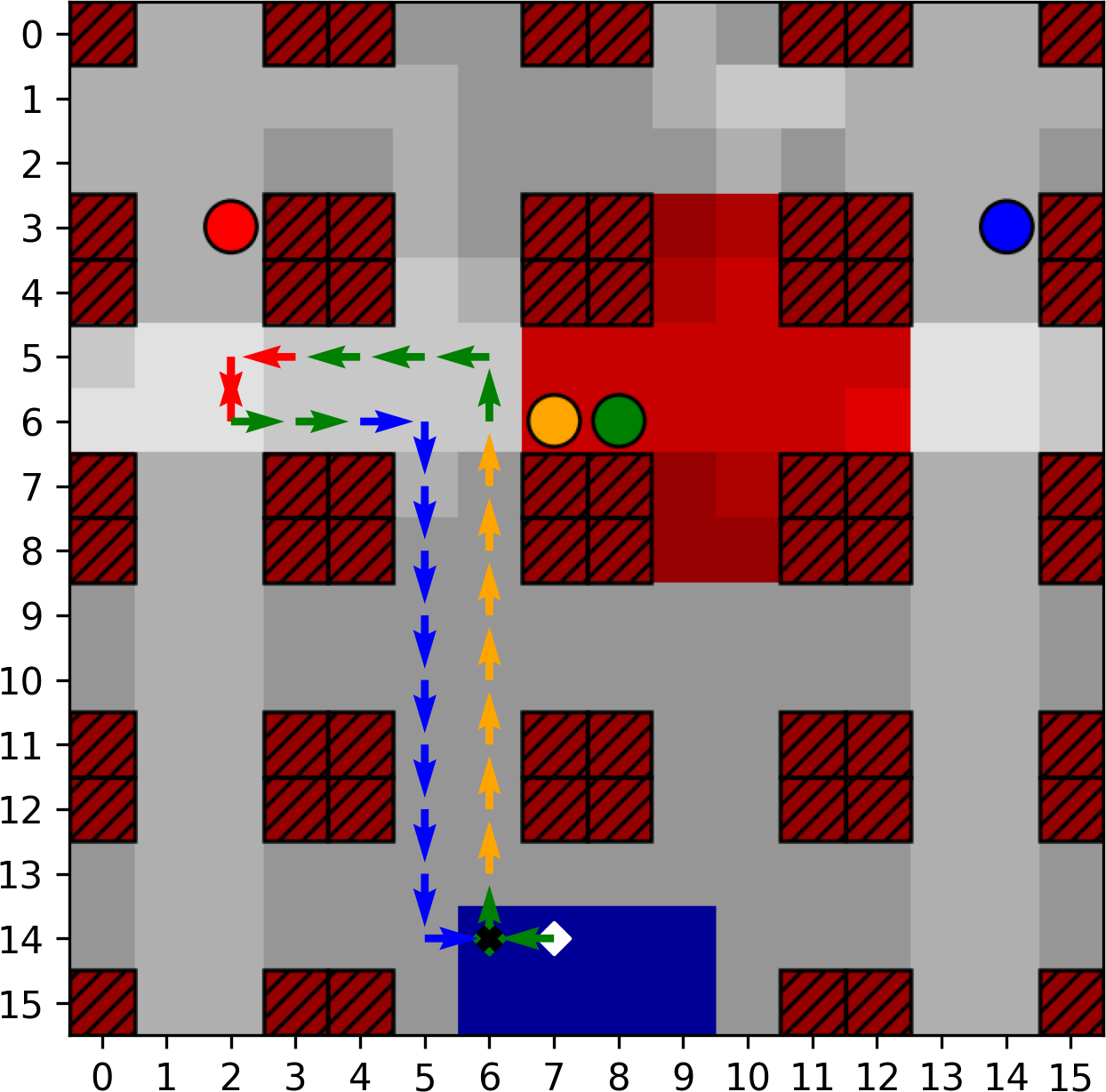}
        \caption{Time 30/35; Data 32.0/50.6}
        \label{fig:mh:c}
    \end{subfigure}\hspace{5pt}%
    \begin{subfigure}{0.47\columnwidth}
        \centering
        \vspace{5pt}
        \includegraphics[width=\textwidth]{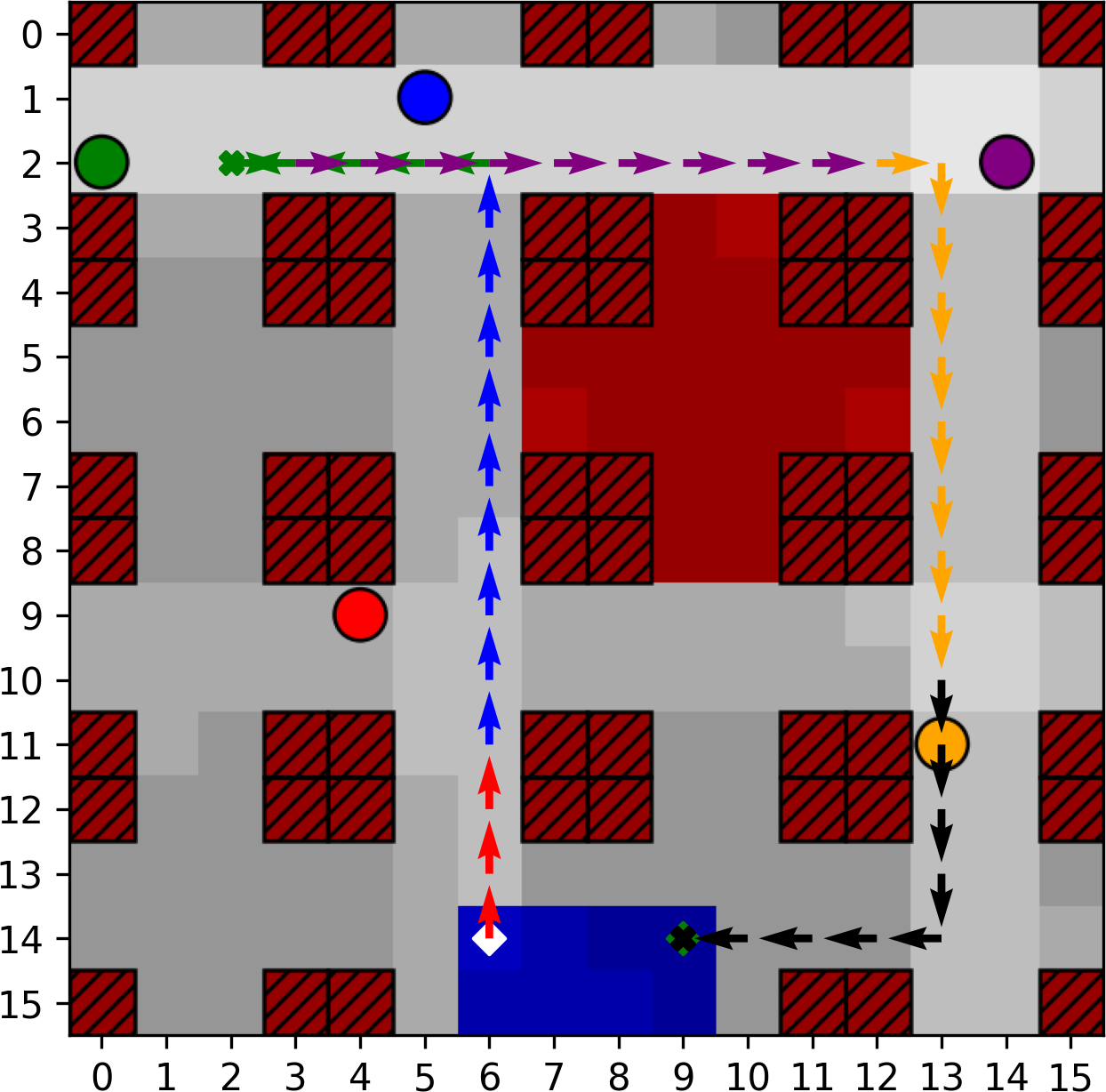}
        \caption{Time 45/65; Data 60.7/60.7}
        \label{fig:mh:d}
    \end{subfigure}
    \caption{Illustration of the same agent adapting to differences in device count and device placement as well as flight time limits, showing used and available flying time and collected and available total data in the Manhattan scenario.}
    \label{fig:mh}
\end{figure}

The main scenario we investigate is defined by a Manhattan-like city structure (see Fig. \ref{fig:mh}) containing regularly distributed city blocks with streets in between, as well as an NFZ district. In this challenging setting, we want to demonstrate the agent's ability to generalize over significant variations in scenario parameters with randomly changing device count ($K \in [2,5]$), device data ($D_0 \in [5.0, 20.0]$ data units), maximum flying time ($b_0 \in [35, 70]$ steps), and eight possible starting positions. Similar to the previous scenario, device positions are randomized throughout the unoccupied map space.

This and the previous scenario are evaluated using Monte Carlo simulations on their full range of scenario parameters with average performance metrics shown in Table \ref{table:metrics}. Both agents show a similarly high successful landing performance. It is expected that the collection ratio must be less than 100\% in some scenario instances depending on the randomly assigned maximum flying time and IoT device parameters.

In Fig. \ref{fig:mh}, four scenario instances chosen from the random Monte Carlo evaluation for device counts of $K \in \{2,3,4,5\}$ for \ref{fig:mh:a} through \ref{fig:mh:d} illustrate the agent's adaptability. With $K=2$ devices in Fig. \ref{fig:mh:a}, finding a trajectory is complicated by the location of the blue device inside the NFZ and the resulting shadowing effects, which have to be deduced by the agent from building and device positions. 
In Fig. \ref{fig:mh:b}, the considerable distance to the red device requires the agent to exhaust its entire flight time. 
For the scenario in Fig. \ref{fig:mh:c} the available flying time $T=35$ is not sufficient to collect all data. Therefore, the agent ignores the isolated blue device and lands early after collecting all data within reach.
In Fig. \ref{fig:mh:d} the agent successfully collects all data in an efficient order while minimizing its flying time, e.g. by turning away from the green device before transmitting all its data. 
We observed that rerunning the same scenario configuration leads to a variation in trajectories which adapt to effects of the random communication channel fading.

\begin{table}[h]
\vspace{-4pt}
\center
\small
\begin{tabular*}{\columnwidth}{ccc}
\toprule[1.5pt]
Metric & Manhattan & Open Field and City\\
\midrule
Has Landed & 99.5\% & 99.9\%\\
Collection Ratio & 94.8\% & 90.0\%\\
Collection Ratio and Landed & 94.6\% & 89.9\%\\
\bottomrule[1.5pt]
\end{tabular*}
\caption{Performance metrics averaged over 1000 random scenario Monte Carlo iterations.}
\label{table:metrics}
\vspace{-10pt}
\end{table}

\section{Conclusion}
\label{sec:conclusion}

We have introduced a new DDQN method with combined experience replay for UAV trajectory planning in an IoT data harvesting scenario. By leveraging a neural network model that exploits information about the environment from centered map layers through convolutional processing, we show that the UAV agent learns to effectively adapt to significant variations in the scenario such as number and position of IoT devices, amount of collectible data or maximum flying time, without the need for expensive retraining or recollection of training data. Using this method, we have shown that the UAV balances the goals of data collection, obstacle avoidance, and minimizing mission time effectively, while not requiring any prior information about the challenging wireless channel characteristics in an urban environment. In future work, we will tackle the issue of scalability to larger maps, namely the linear increase of trainable parameters in the flatten layer with map area. We also envision to combine our approach with multi-task reinforcement learning or transfer learning \cite{DulacArnold2019}, as well as extending the UAV's action space to altitude control.

\bibliography{bib_globecom}
\bibliographystyle{ieeetr}

\end{document}